\documentclass[10pt,twocolumn,letterpaper]{article}

\usepackage{cvpr}
\usepackage{times}
\usepackage{epsfig}
\usepackage{graphicx}
\usepackage{amsmath}
\usepackage{amssymb}
\usepackage{subcaption}
\usepackage{xcolor}
\usepackage{tabularx} 
\usepackage{float} 

\newcolumntype{L}[1]{>{\raggedright\arraybackslash}p{#1}}
\newcolumntype{C}[1]{>{\centering\arraybackslash}p{#1}}
\newcolumntype{R}[1]{>{\raggedleft\arraybackslash}p{#1}}

\usepackage[pagebackref=true,breaklinks=true,letterpaper=true,colorlinks,bookmarks=false]{hyperref}

\cvprfinalcopy 

\def\cvprPaperID{9} 

\ifcvprfinal\fi
\begin{document}

\title{Detection of Single Grapevine Berries in Images Using Fully Convolutional Neural Networks}


\author{Laura Zabawa$^{1*} $, Anna Kicherer$^2$, Lasse Klingbeil$^1$, Andres Milioto$^1$, \\
Reinhard T\"opfer$^2$, Heiner Kuhlmann$^1$, Ribana Roscher$^1$\\
$^1$Rheinische Friedrich-Wilhelms-Universit\"at Bonn, Institute of Geodesy and Geoinformation\\
\and
$^2$Julius K\"uhn-Institut, Federal Research Centre of Cultivated Plants, \\
Institute for Grapevine Breeding Geilweilerhof\\
{\tt\small $^*$zabawa@igg.uni-bonn.de}
}

\maketitle

\begin{abstract}
Yield estimation and forecasting are of special interest in the field of grapevine breeding and viticulture.
The number of harvested berries per plant is strongly correlated with the resulting quality. Therefore, early yield forecasting can enable a focused thinning of berries to ensure a high quality end product. 
Traditionally yield estimation is done by extrapolating from a small sample size and by utilizing historic data. 
Moreover, it needs to be carried out by skilled experts with much experience in this field.  
Berry detection in images offers a cheap, fast and non-invasive alternative to the otherwise time-consuming and subjective on-site analysis by experts. 
We apply fully convolutional neural networks on images acquired with the Phenoliner, a field phenotyping platform. We count single berries in images to avoid the error-prone detection of grapevine clusters. 
Clusters are often overlapping and can vary a lot in the size which makes the reliable detection of them difficult.
We address especially the detection of white grapes directly in the vineyard. 
The detection of single berries is formulated as a classification task with three classes, namely 'berry', 'edge' and 'background'. 
A connected component algorithm is applied to determine the number of berries in one image. 
We compare the automatically counted number of berries with the manually detected berries in 60 images showing Riesling plants in vertical shoot positioned trellis (VSP) and semi minimal pruned hedges (SMPH). 
We are able to detect berries correctly within the VSP system with an accuracy of 94.0 \% and for the SMPH system with 85.6 \%. 
\end{abstract}

\section{Introduction}
Wine represents one of the oldest and economically most important fruit crops. 
The breeding of new and robust varieties is therefore of special interest to the community. 
Wine breeders do not only need information about the health and growth status of the plants, they also call for early yield forecasting. 

\begin{figure}[t]
\begin{center}
   \includegraphics[width=0.9\linewidth]{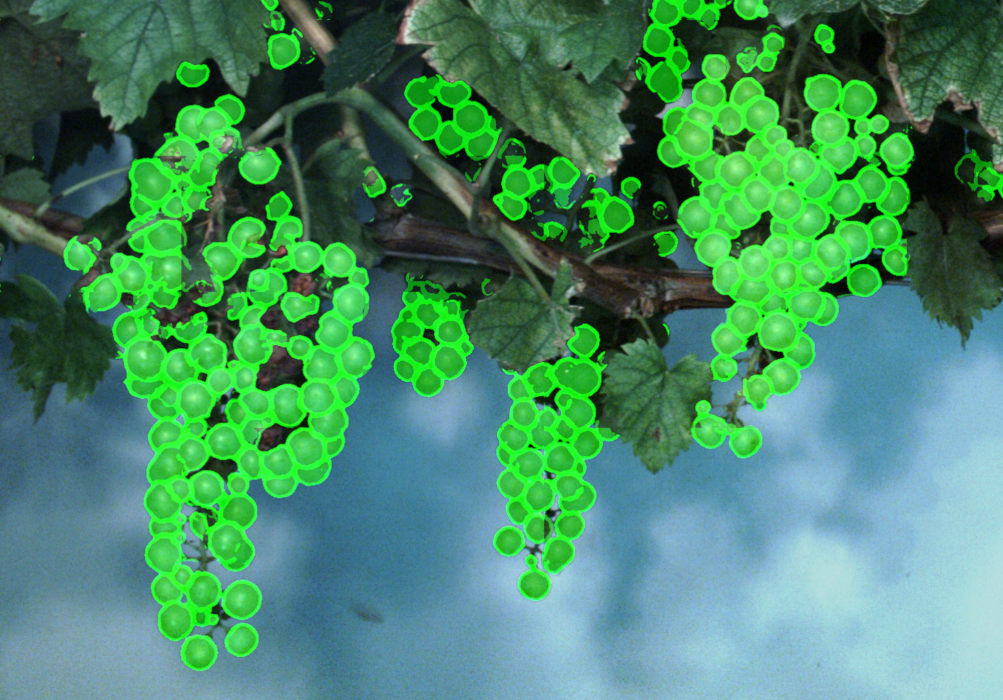}
\end{center}
   \caption{Example for recognized berries. In order to ensure the detection of single berries, each berry is surrounded by pixels classified as 'edge'.}
\label{fig:Classes}
\end{figure}

Phenotypic traits can give information about the health and growth status but are difficult to acquire on large scales. This can be explained with the perennial nature of wine which requires an on-site analysis by skilled experts. This leads to expensive, subjective, and labour-intensive results. 
The application of the BBCH Scale (Scale of the Biologische Bundesanstalt, Bundessortenamt und CHemische Industrie) \cite{Bloesch08} or the OIV (Organisation Internationale de la Vigne et du Vin) descriptor \cite{OIV01}, which mainly relied on subjective visual interpretation, are the common approach.

Yield estimation by manual and visual means is time consuming and error prone as well. 
Traditionally it is done by counting and weighing grape bunches from a selected number of vines and extrapolating these numbers to the whole wine plot. 
The accuracy increases the closer the measurements are acquired to the harvest date.
Early yield estimation however offers the possibility to take early actions. One example are guided and objective thinning procedures to avoid an overload of the plants, which would lead to decreased berry quality. 

\begin{figure}[t]
\begin{center}
   \includegraphics[width=0.9\linewidth]{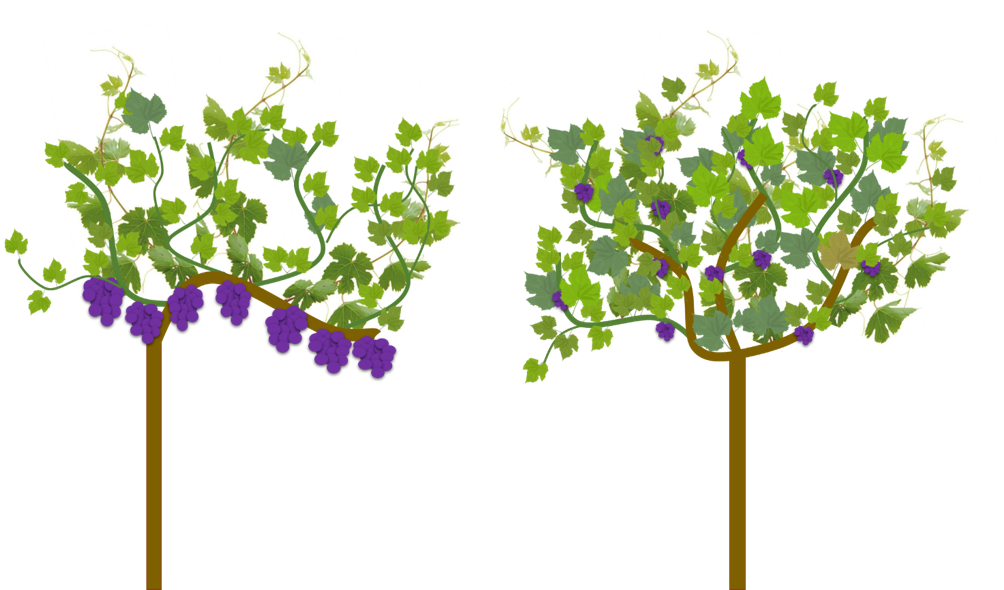}
\end{center}
   \caption{Examples of the different training systems. On the left is an example of a plant in a vertical shoot positioned system (VSP). It features one main branch which grows over multiple years. The other branches regrow every year. The grapes are mainly positioned on the bottom of the canopy and are often not much occluded. The right side shows a plant which is trained in a semi minimal pruned hedge (SMPH). It features more branches and leafs. The grapes grow all over the plant and are often occluded by canopy.}
\label{fig:SMPH}
\end{figure}

Furthermore, different training systems pose different difficulties for the yield estimation. The traditional vertical shoot positioned (VSP) training system features compact bunches and only few overlapping leafs. This is due to the fact that only one main branch exists and many leaves are cut down. Most grapes grow on the bottom part of the canopy.
The new training system semi minimal pruned hedges (SMPH) features a looser bunch structure and an increased leaf coverage. They have many branches and a thick layer of canopy. Berries can occur in all places of the canopy.
Both are challenging regarding detection and quantification of berries. For the SMPH the yield estimation with the number of berries might not be sufficient due to the in-homogeneous size of the berries. Therefore we aim in the long run not only for the number of berries but also for an investigation of the berry size.

A reliable and accurate yield estimation demands an objective and fast high-throughput on-site investigation. The number of bunches and grapes is heavily correlated to the resulting yield. The traditional yield estimation is based on the investigation of bunches. This is a hard task by visual means only because bunches are hard to distinguish from each other. 
One way is the automatic detection of single berries in images which offers a cheap and fast alternative to manual expert-based procedures. 
We present a robust detection pipeline which detects single berries in images by defining a three-class classification task and the application of a fully convolutional neural network. Furthermore our approach enables the investigation of the berry size which might be helpfull for an even better yield estimation.

Our main contribution is the reformulation of an instance segmentation as an semantic segmentation task. 
For this, we define the classes 'berry', 'edge' and 'background', where the use of the classes 'berry' and 'edge' enables the differentiation between single berries within a cluster. This definition enables not only the counting and location identification of each berry but also the possible investigation of another important phenotypic trait, the size.

\section{Related Work}
For many plants such as perennial crops it is important to investigate the phenotypic traits on-site and to acquire large scale data sets to avoid extrapolating from small sample sizes. 
Image processing and machine learning enable high-throughput phenotyping which is important for this task. 
The main advantages of automatic procedures are objectivity, repeatability and high quality. 

Two main approaches exist for the problem of counting in images. The first one is a regression-based while the second one relies on detection. 
One of the earliest works which skips the detection while having counting in mind was done by Lempitsky and Zisserman \cite{Lemptisky10}. They utilized dot annotated training data to learn an estimation for image density by minimizing a regularized risk quadratic cost function. The integration of the image density leads to the count of objects in the image.
Later, Xie et al. \cite{Xie16} used convolutional neural networks to regress the spatial density over images. Their application was focused on microscopical images of cells.
Arteta et al. \cite{Arteta16} tackled the counting of penguins which stated a complex real word problem. The data set featured object occlusions and scale variations. They used a deep multitask architecture to combine a density estimation with a foreground-background separation and local uncertainty estimations. 
The count-ception network was introduced by Cohen et al. \cite{CohenLB17}. They applied regression in a fully convolutional way to the images and counted redundantly and average all predictions.
A first class agnostic approach was presented by Lu et al. \cite{Lu18}. They reformulated the counting as an matching problem which enabled the use of video annotations as training data. By using an adapter module they are able to easily extend their approach for arbitrary counting tasks.
All regression methods focus on avoiding the detection task and offer at best a spatial information additional to the count. 
In contrary to this, we provide a framework which is able to provide more comprehensive and valuable phenotypic traits such as the berry size.

An overview about agricultural applications was done, for example, by Gongal et al. \cite{Gongal15}, in which they compare different sensors and algorithms for fruit detection and localization with a robotic background. 

Some research focuses on simple image analysis frameworks by detecting geometric objects in images. 
Nyarko et al. \cite{Nyarko18} detect fruits in uncontrolled conditions by detecting convex surfaces.
The surfaces are classified with a k-nearest neighbour approach into 'fruit' and 'not fruit'. 
They performed experiments on 4 different fruits including tomato, nectarine, pear and plum.
Roscher et al. \cite{Roscher13} investigate berry sizes in images taken by a consumer camera using an image analysis framework. 
They apply a circular Hough-Transform to the images and classify the results into 'berry' and 'not berry' using a Markov random field. 
However, the counting of grapes is not realized.
The first large scale experiment was presented by Nuske et al. \cite{Nuske14} in 2014. They evaluated their system in a realistic experimental setup over several years and hundreds of vines. 
Nuske et al. use a camera system with illumination which is mounted on a vehicle. 
A circular Hough-transform is applied to the images, and resulting berry candidates are classified by texture, colour and shape features. 
Neighbouring berries are grouped into clusters.

Most image analysis methods rely on carefully chosen features and thresholds which require finetuning for every application.
Since 2012 the application of neural networks to image classification problems became common, after Krizhevsky \cite{Krizhevsky12} won the most popular image classification challenge. 
In 2015 fully convolutional neural networks were proposed by Long et al. \cite{Long15} which enable a pixelwise semantic segmentation of images. 
The pixelwise identification of single objects, so-called instance segmentation, is another major development of the deep learning community. 
He et al. \cite{He17} proposed Mask-RCNN, a major development in this research area.

Neural networks have also been applied for the analysis of grapevine images.
In 2016, Aquino et al. \cite{Aquino16} presented a smartphone application which was able to recognize and count green berries in images.
They put a black box around single clusters and took an image. 
They applied a circular light reflection detection and classified the results with a neural net. 
In a later approach, they implemented a more automated data collection approach and discarded the need for a black box around the grape cluster \cite{Aquino17}.
Rudolph et al. \cite{Rudolph18} concentrated on the detection and quantization of grapevine inflorescences in images. 
They first identified image regions containing inflorescences with a neural network and applied a circular Hough-transform on the resulting image regions.

In contrast to other multi-step approaches, we present an end-to-end detection framework for berries. 
The potentially high number of berries in images (up to 1000) is challenging and the detection cannot be accomplished easily with commonly used instance segmentation approaches. 
For example, many of those approaches reduce the runtime by limiting the number of instance proposals. 
We achieve a proposal-less detection and segmentation of single berries.
Therefore, we choose to formulate the berry detection task as a semantic segmentation which is able to divide single berries by introducing an additional class 'edge'. This proceeding allows later the extraction of further phenotypic data like the berry size.

\section{Data}
\begin{figure}[t]
\begin{center}
   \includegraphics[width=0.8\linewidth]{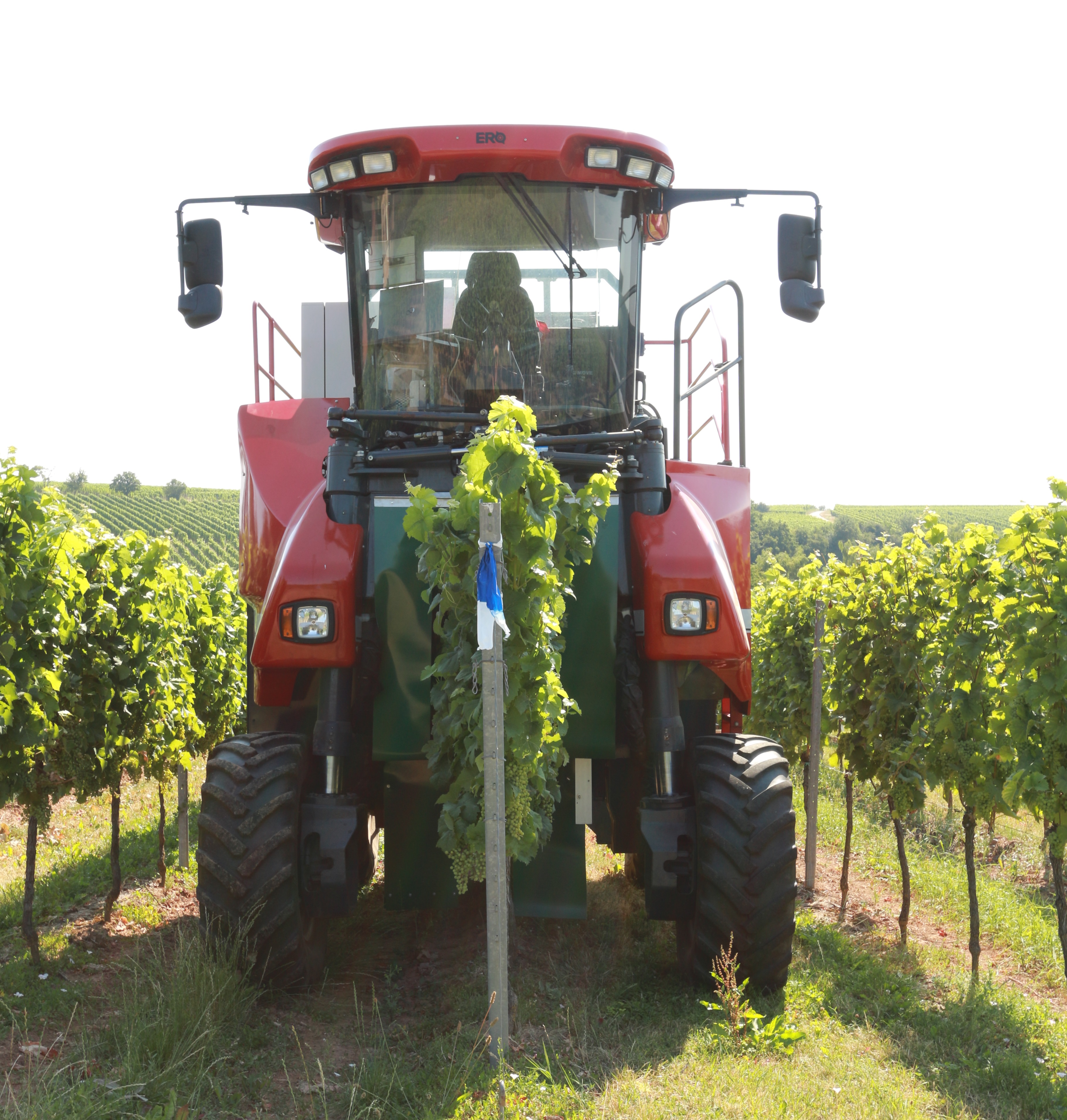}
\end{center}
   \caption{Field phenotyping platform called Phenoliner \cite{Kicherer17}. The Phenoliner is based on a grapevine harvester. The harvesting equipment is replaced by a calibrated stereo camera system. The camera system produces overlapping images of the plants covering approximately $1.2$~m vertically of the wine row.}
\label{fig:Phenoliner}
\end{figure}

Images were acquired with the Phenoliner \cite{Kicherer17}, a field phenotyping platform which is shown in Fig. \ref{fig:Phenoliner}. 
The Phenoliner is composed of a modified grapevine harvester. The harvesting equipment was removed and a camera system with several lamps was installed. The camera system consists of three vertically aligned cameras. The Phenoliner is able to cover $1.2$~m of the plants vertically. Each image has $2048 \times 2592$ pixels.

Data were acquired on three different dates in 2018. The first images were taken shortly before the thinning, the second after the thinning and the last one a few days before the harvest.

The measurements were taken in different experimental vineyard plots at the JKI Geilweilerhof located in Siebeldingen, Germany. Three different varieties were observed: Regent, Riesling and Felicia. 

The observation of every variety contained at least two different training systems, semi minimal pruning hedge (SMPH) and vertical shoot positioning (VSP), illustrated in Fig. \ref{fig:Phenoliner}. Both systems feature different difficulties. 
The leaf area is drastically reduced in the VSP and berries are well visible. The grape bunches are compact and berries are mostly clustered in them. The plant itself has one major branch which grows over several years while other branches and canopy is cut down every year.
In the SMPH many berries are occluded by leafs due to the minimal pruning. Furthermore they feature a loose bunch structure and single small bunches appear between a lot of leaf coverage. The plant features a few main branches which leads to the extended canopy cover.

Images showing the VSP feature up to 890 berries per image, with an average of 329 berries. For the SMPH up to 1.106 berries are shown per image. The average number of berries is 556 berries.

\subsection{Image Annotation}
\label{sec:annotation}
The detection of single berries is formulated as a semantic segmentation task with three classes: 'berry', 'edge', and 'background'. 
Pixels of the class 'berry' are surrounded by pixels of the class 'edge', while remaining pixels are assigned to the class 'background'.

We manually annotated 32 images, where each berry is colored independently. This means that every berry is colored in one out of four colours and touching berries never have the same colour to ensure the identification of every single berry.
For every individually coloured berry an edge is computed with a fixed size. This edge is labeled as 'edge' and the rest is labeled as 'berry'. Every pixel which was not annotated manually is uniformly labeled as 'background'.

To evaluate the counting we annotated 60 images of Riesling with dot annotations. This means that every berry in the image was marked with a single dot to give an objective reasoning about the berry count and spatial allocation. 30 of these images show plants in the VSP, the other 30 plants in the SMPH.

\section{Methods}
\subsection{Overall Workflow}
\label{sec:work}
\begin{figure*}[t]
    \centering
    \begin{subfigure}[b]{0.45\textwidth}
        \centering
        \includegraphics[width=\textwidth]{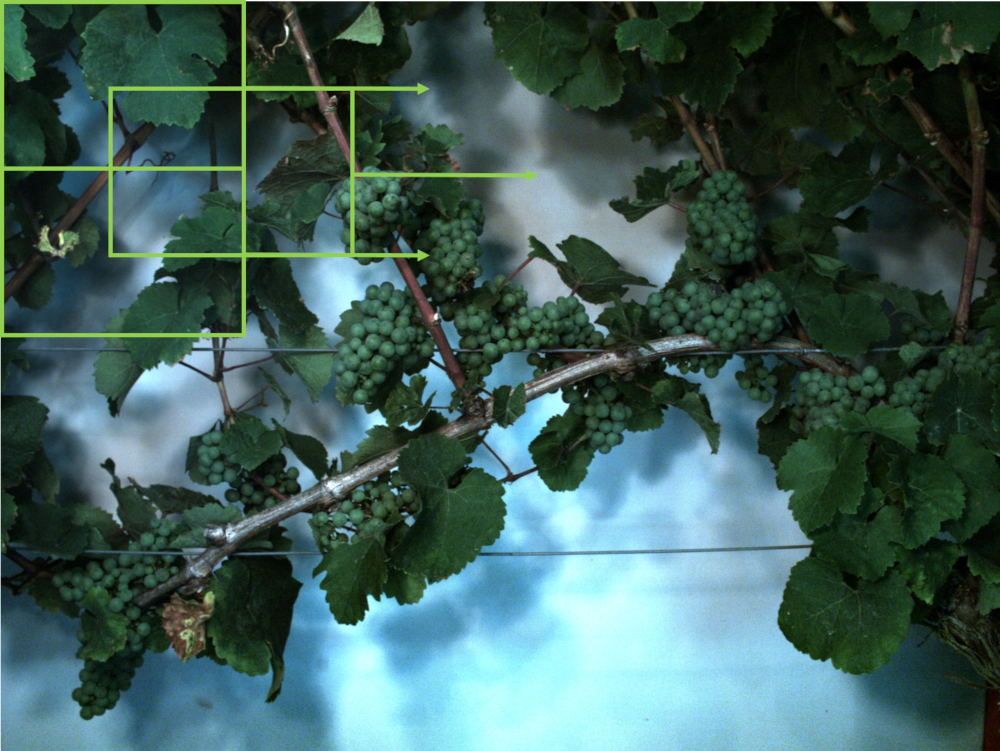}
        \caption[Overlapping windows are slid over the original image and extract small image patches. Each image patch is classified with the neural network.]%
        {{\small Original image}}    
        \label{fig:Orig}
    \end{subfigure}
    \quad
    \begin{subfigure}[b]{0.45\textwidth}  
        \centering 
        \includegraphics[width=\textwidth]{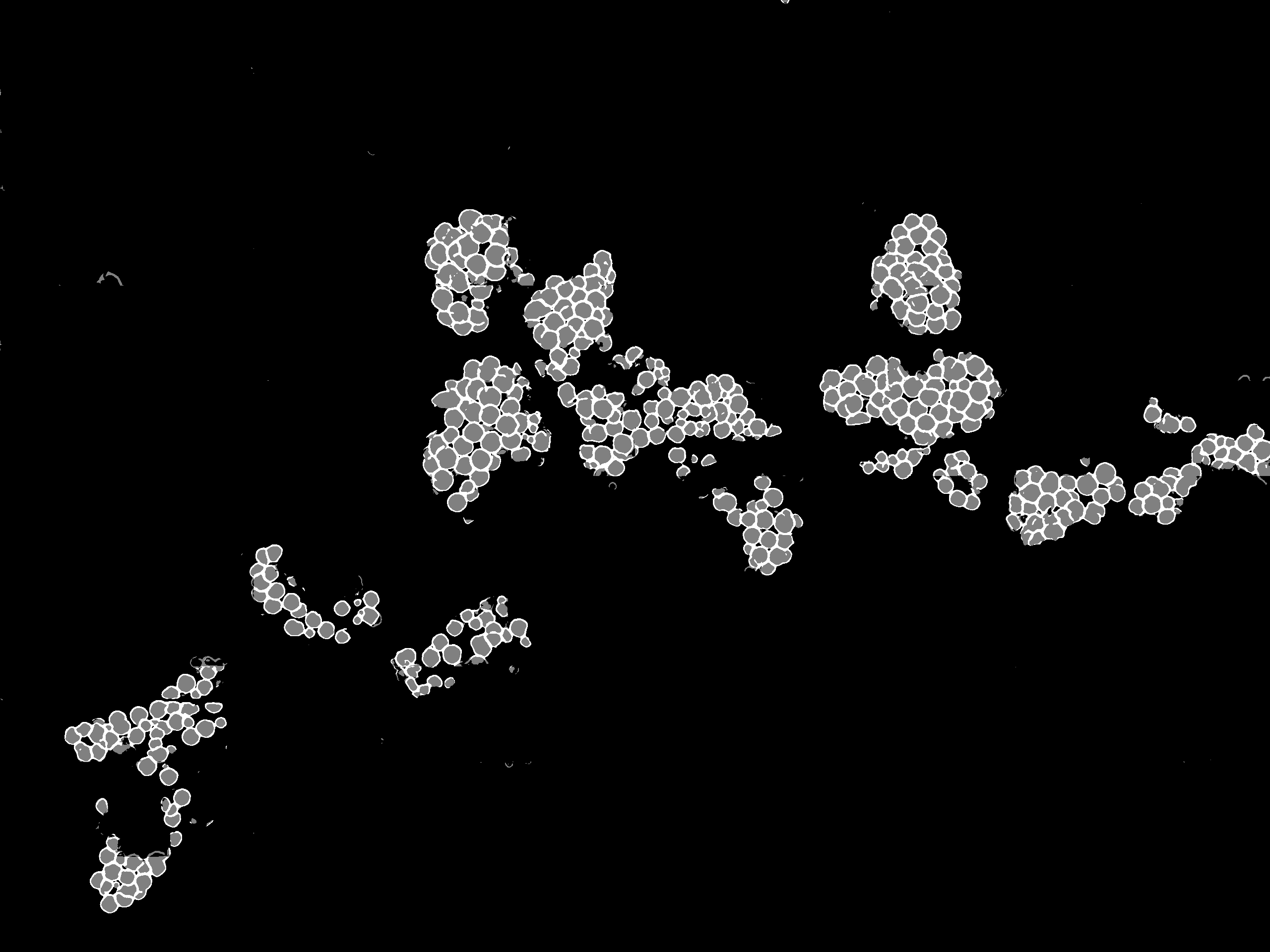}
        \caption[The classified image patches are used to reconstruct a mask for the original image using a majority vote.]%
        {{\small Prediction mask}}    
        \label{fig:Mask}
    \end{subfigure}
    \vskip\baselineskip
    \begin{subfigure}[b]{0.45\textwidth}   
        \centering 
        \includegraphics[width=\textwidth]{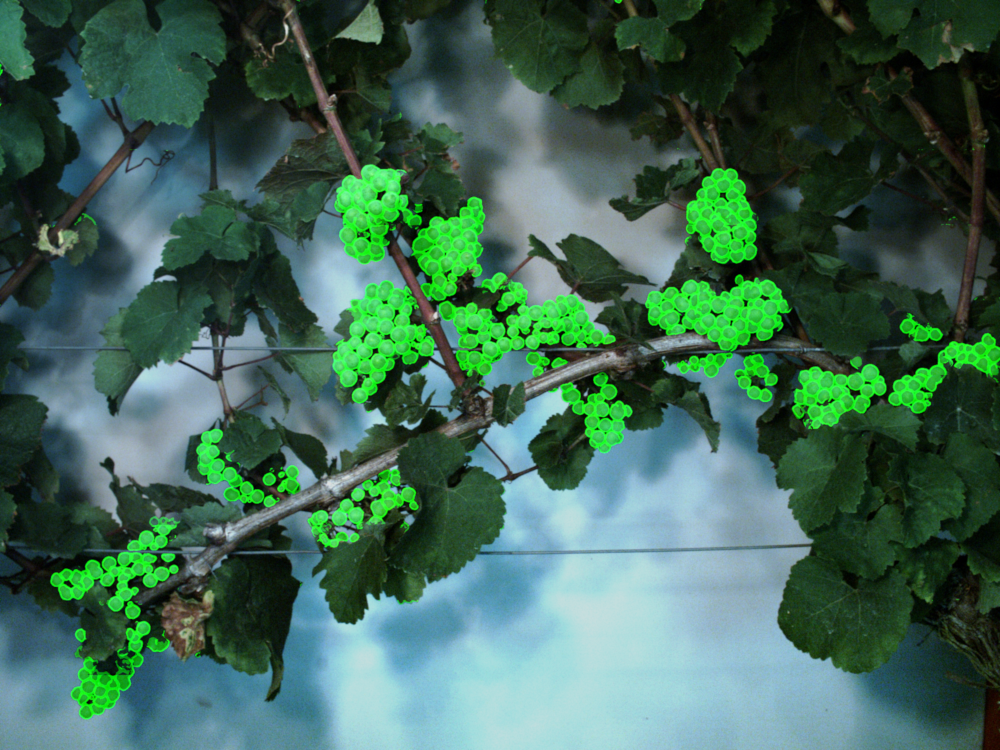}
        \caption[]%
        {{\small Overlayed Prediction}}    
        \label{fig:predi}
    \end{subfigure}
    \quad
    \begin{subfigure}[b]{0.45\textwidth}   
        \centering 
        \includegraphics[width=\textwidth]{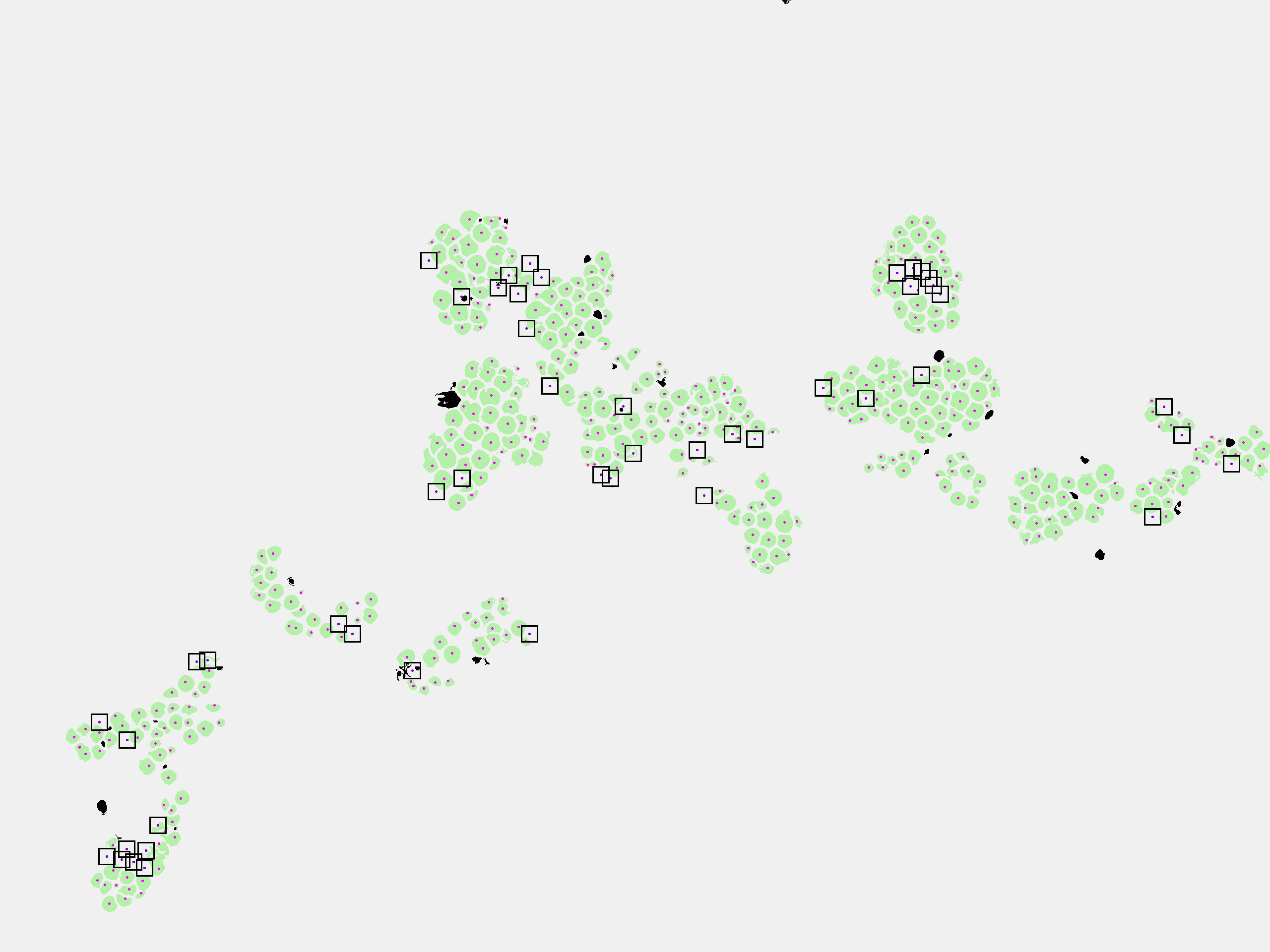}
        \caption[]%
        {{\small Evaluation}}    
        \label{fig:eval}
    \end{subfigure}
    \caption[ The average and standard deviation of critical parameters ]
    {\small Classification workflow. (a) overlapping windows are slid over the original image and extract small image patches. Each image patch is classified with the neural network. (b) the classified image patches are used to reconstruct a mask for the original image using a majority vote. (c) the mask overlaps the original image which allows a visual evaluation. (d) evaluation with manually counted berries. Green regions show correctly classified components which include a manual marker. Black components show incorrectly classified regions. Black dots with a box around show manually counted berries which were not found by the neural network.} 
    \label{fig:workflow}
\end{figure*}

Fig. \ref{fig:workflow} illustrates the workflow for the application on a test image. 
In Fig. \ref{fig:Orig}, the image is cut into overlapping image patches with a sliding window approach, where each patch is classified separately with the neural network. 
A mask for the original image is reconstructed from the patches (Fig. \ref{fig:Mask}). 
The resulting class for each pixel is determined by a majority vote over all overlapping image patches. 

A connected component algorithm is used to determine the number of berry components. The edges and the background are not considered. We discard objects which are likely noise fragments by thresholding with a minimum number of 25 pixels for each component. 

A visual inspection is possible when the resulting prediction mask is overlayed with the original image, as shown in Fig. \ref{fig:predi}. 
The evaluation is realized by comparing the prediction mask with manually marked berries in the images (Fig. \ref{fig:eval}). 
Green components are correctly identified berries. They overlap with the manual markers. Black regions are incorrectly classified as berry. 
Black dots with boxes show berries which were not detected by the neural network.

\subsection{Semantic Segmentation Network}

The main contribution of this paper is the reformulation of the counting and instance segmentation task as a semantic segmentation problem. This turns the problem of detecting and segmenting each individual berry instance in the image into a pixel-wise classification of~\ref{fig:Orig} into the classes 'berry', 'edge', and 'background' resulting in an inferred mask like the one in Fig.~\ref{fig:eval}.

The images that the collection platform provides are DSLR-quality, which makes them really expensive to work with in terms of memory consumption. This becomes a problem in moving platforms that carry on-board processing hardware. The two possible ways to make the processing more efficient using convolutional neural networks (CNNs) are to downsample the images, or to work with image patches. Because the berries are small, the downsampling process hurts performance, resulting in missing berries and wrong classification results. Instead, working with overlapping patches allows us to obtain a classification mask for all the pixels in the original full-resolution image, without hurting performance. However, this means that we need to run the CNN multiple times, and this requires a light-weight, yet powerful neural network in order to efficiently process each patch.

For this, we use an hourglass encoder-decoder architecture based on the inverted residual concept introduced in~\cite{Sandler18}. The encoder backbone used is MobileNetV2~\cite{Sandler18} which is an efficient and lightweight feature extractor formulation for mobile applications that achieves results close to the state-of-the-art for tasks such as classification, segmentation, and detection. The decoder used is DeepLabV3~\cite{Chen18}. This combination results in a fully convolutional semantic segmentation CNN that can accurately resolve the proxy task of segmenting berries, edges, and background, while still performing fast in the moving platform.

\subsection{Post-processing}
In order to reduce the amount of components which are misclassified as 'berry', we filter our results by assuming general geometric properties of berries. 
Roundness is the main property of berries.
We explore two possibilities to remove objects that do not feature the definition of roundness. 
After identifying connected components we compute properties for every single component, where minor and major axis as well as the area are the most interesting ones for us.
By comparing the minor and major axis we remove objects which have a relation between both under 0.3. 
By using the mean of the minor and major axis we compute a radius. With this radius we compute the area which the component should have if it was a circle. Components with an actual area smaller than 0.3 times the computed area are disregarded as well.
Furthermore we investigate the edges around each detected berry. Components which are classified with a high confidence are often surrounded completely by an edge. Therefore we don't consider components which are surrounded by less than 40\% by an edge. 

\section{Experiments}
\subsection{Experimental Setup}
We perform the following experiments:
\begin{itemize}
    \item A variation of edge thickness values and its influence on the test accuracy, 
    \item analysis of intersection over union (IoU),
    \item an analysis of the influence of post-processing, and
    \item the investigation of the berry counting
\end{itemize}

The network is trained on overlapping image patches extracted from the 32 edge annotated images. Each patch has $384 \times 512$ pixels and 50 \% overlap in vertical and horizontal direction. This reduces the training time drastically compared to training on the full resolution of the images. 
We perform three kind of data augmentations.
The first augmentation is horizontally flipping the original image. The second one is blurring the image with a random kernel size between $3$ and $7$. The last augmentation is gamma shifting randomly between $0.8$ and $1.2$.
In the end, we get roughly $4.900$ image patches from 32 annotated images. The training set contains $90\%$ of these patches and the test set the remaining $10\%$. To evaluate the classification accuracy we investigate the intersection over union (IoU). The IoU describes the relation between area of overlap between ground truth and prediction and the area of union between both. 

\begin{table}[t]
\begin{center}
\begin{tabular}{|L{1.7cm}|C{1cm}|C{2cm}|C{2cm}|}
\hline
Training System & Edge [pixel] & Correct Detection [\%] & Misclassified [\%]\\
\hline\hline
VSP & 2 & 93.9 & 19.2\\
VSP & 3 & 92.6 & 25.4\\
SMPH & 2 & 89.0 & 25.6\\
SMPH & 3 & 85.3 & 28.4\\
\hline
\end{tabular}
\end{center}
\caption{Comparison of various edge thickness values on different training systems. 'Correct detection' indicates the percentage of manually counted berries which lie within a berry component. 'Misclassified' is the percentage of components which are not overlapped with a manually counted berry.}
\label{tab:Edge}
\end{table}

Furthermore we evaluate our network on 60 images of Riesling, as described in Sec. \ref{sec:annotation}. With these images we focus on the evaluation of the total count of berries in images patches. The images are therefore cut into the same dimensions as the patches which are used for training. This means we have 750 image patches for each training system where we compare the actual number of berries with the detected number of berries. 
Half of the images show plants in the VSP and the other half in SMPH. 
We manually marked the center of all berries in each image as reference with dots.
A berry is correctly detected if the marker lies within the berry segment. 
Finally, we compare the detected berries with the manually marked berries.

The learning rate for the network is $0.001$ and the momentum is $0.9$. The learning rate is decreased by $0.99$ after 5 epochs.

\subsection{Results}
Tab. \ref{tab:Edge} shows that with our main setup we are able to detect between $85 - 94 \%$ of manually identified berries. 
However, up to $28 \%$ of detected components are  incorrectly classified as berry. 
Therefore further experiments with filtering out misclassified objects are done in Sec. \ref{sec:filter}. 

\subsubsection{Edge Thickness and Training System}
The correct detection of single berries depends on the accuracy of the class 'edge'. 
We tested two different edge thickness values of 2 and 3 pixels and evaluated if a stable and explicit differentiation between single berries can be ensured.
We apply our classifier with different edge thickness values on two different training systems: VSP and SMPH.

Tab. \ref{tab:Edge} shows that we recognize more berries correctly in the VSP compared to the SMPH. 
That means that the influence of the different edge thickness values on the number of correct detections is smaller for VSP than for SMPH. 
For SMPH the smaller edge thickness of 2 pixels correctly identifies 4\% more berries. 
The SMPH is characterized by inhomogeneous berry sizes with a higher number of small berries than VSP.
Smaller berries might only consist of edge pixels, and therefore, cannot be detected by our proposed method.

Furthermore we investigate the intersection over union (IoU) for every class that can be seen in Tab. \ref{tab:Acc}. 
For both training systems we achieve an IoU of over 75~\%. The data set with an edge width of 2 pixels has a slightly better IoU for the class 'berry', but shows a notably worse accuracy for class edge with 41.7~\%. 
However, a thin edge is hard to reproduce and therefore the result is reasonable.

\begin{table}[t!]
\begin{center}
\begin{tabular}{|l|c|c|c|c|c|}
\hline
& Edge [3 pixels] & Edge [2 pixels] \\
\hline\hline
Average IoU [\%]& 76.0 & 72.2 \\
IoU Background [\%]& 99.0 & 98.9 \\
IoU Berry [\%]& 75.3 & 76.0 \\
IoU Egde [\%]& 53.7 & 41.7 \\
\hline
\end{tabular}
\end{center}
\caption{Investigation of the intersection over union (IoU). The IoU is better for berry than edge which can be explained with the nature of the classes. The overlay of thin edges is unlikely.}
\label{tab:Acc}
\end{table}

\subsubsection{Filtering} 
\label{sec:filter}

\begin{figure}[t]
\begin{center}
   \includegraphics[width=1\linewidth]{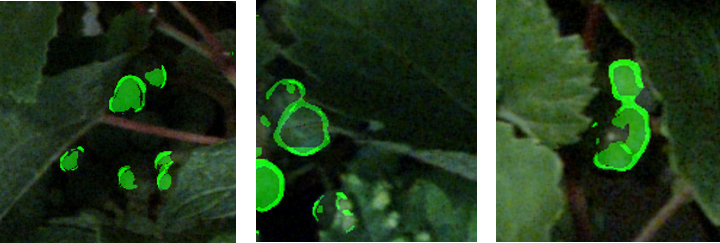}
\end{center}
   \caption{Incorrectly segmented objects. The examples show objects which are not sufficiently round (left and right) and in many cases the berries are not completely surrounded by an edge.}
\label{fig:Filter}
\end{figure}

Fig. \ref{fig:Filter} shows examples for objects which are not round. On the left and right we see dark image regions which are incorrectly classified as berries. The objects are not sufficiently round and not surrounded correctly by an edge. In the middle, the component itself looks like an edge and is not round as well. 

\begin{table*}[t]
\begin{center}
\begin{tabular}{|l|c|c|c|c|c|}
\hline
Training System & Axis & Area & Edge & Correct Detection [\%] & Misclassified [\%]\\
\hline\hline
VSP & - & - & - & 93.9 & 19.2\\
VSP & 0.3 & - & - & 93.7 & 14.9\\
VSP & 0.3 & 0.3 & - & 93.7 & 14.1\\
VSP & 0.3 & 0.3 & 0.4 & 92.5 & 10.3\\
SMPH & - & - & - & 89.0 & 25.6\\
SMPH & 0.3 & - & - & 88.9 & 22.0\\
SMPH & 0.3 & 0.3 & - & 88.8 & 20.9\\
SMPH & 0.3 & 0.3 & 0.4 & 86.9 & 14.7\\
\hline
\end{tabular}
\end{center}
\caption{Comparison of different filter strategies. Axis means that the relation between the minor and major axis of each component isn't allowed to be smaller than 0.3. For the computation of a circle area we compute the radius of each component as the mean of the minor and major axis. We then compare the computed area with the actual area of each component. The actual area is not allowed to be smaller than 0.3 times the circle area. Edge means that every component needs to be surrounded by at least $40 \%$ of edge.}
\label{tab:Filter}
\end{table*}

Tab. \ref{tab:Filter} shows the results after applying different filter methods. The application of the simple geometric prior information reduces the number of misclassifications by around 4\%. After incorporating the edge criterion we are able to reduce the misclassifications to 10\% and 15\% while only losing around 1\% of the correctly classified berries. 



\subsubsection{Berry Count}
The evaluation of the berry counts is done on 30 independent, dot annotated images of Riesling in the VSP and 30 in the SMPH. 

\begin{figure*}[t]
    \centering
    \begin{subfigure}[b]{0.48\textwidth}
        \centering
        \includegraphics[width=\textwidth]{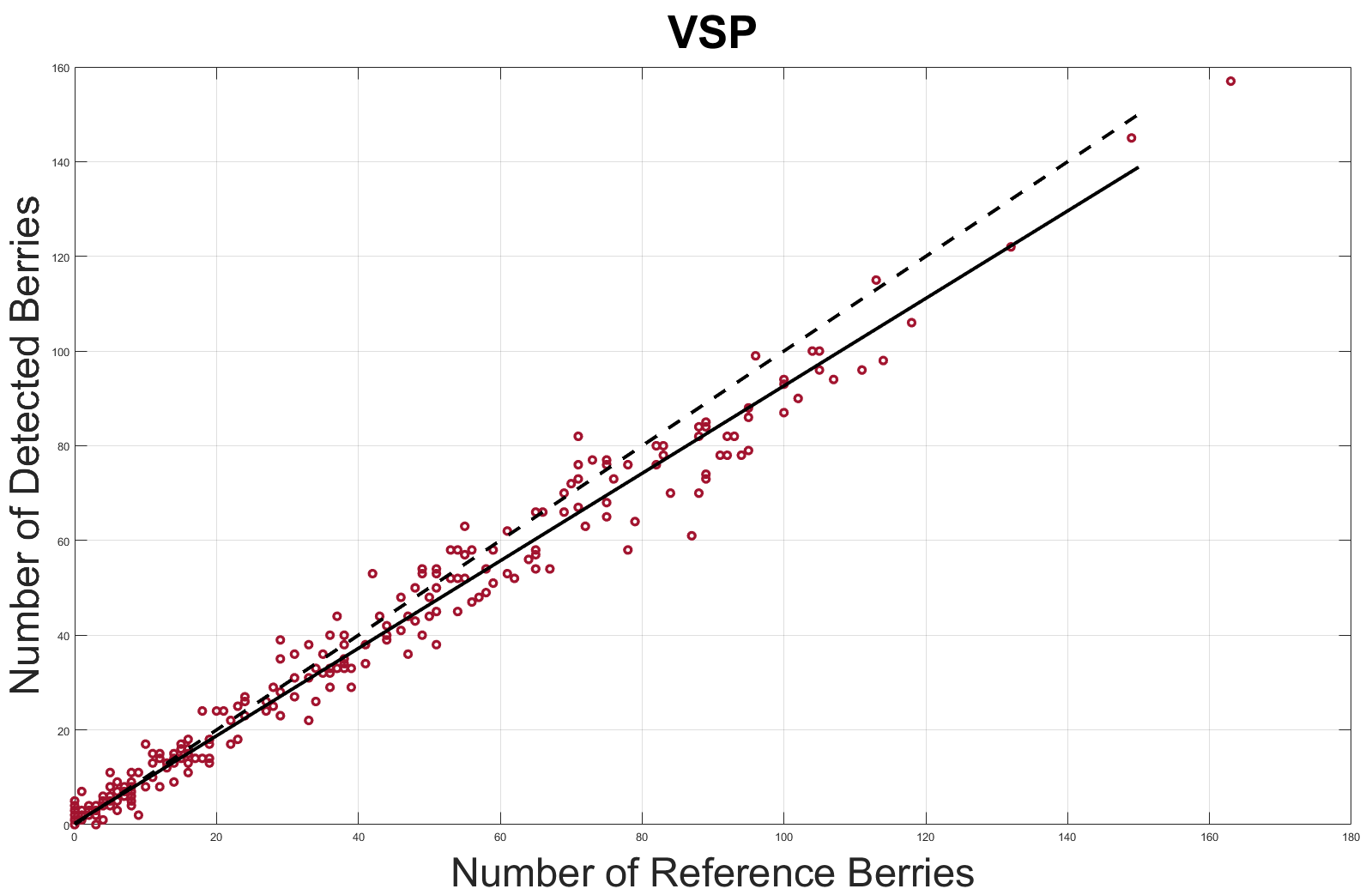}
        \caption[Overlapping windows are slid over the original image and extract small image patches. Each image patch is classified with the neural network.]%
        {{\small $R^2$-Plot for the VSP with $R^2 = 88.33$}}    
        \label{fig:R2_VSP}
    \end{subfigure}
    \quad
    \begin{subfigure}[b]{0.48\textwidth}  
        \centering 
        \includegraphics[width=\textwidth]{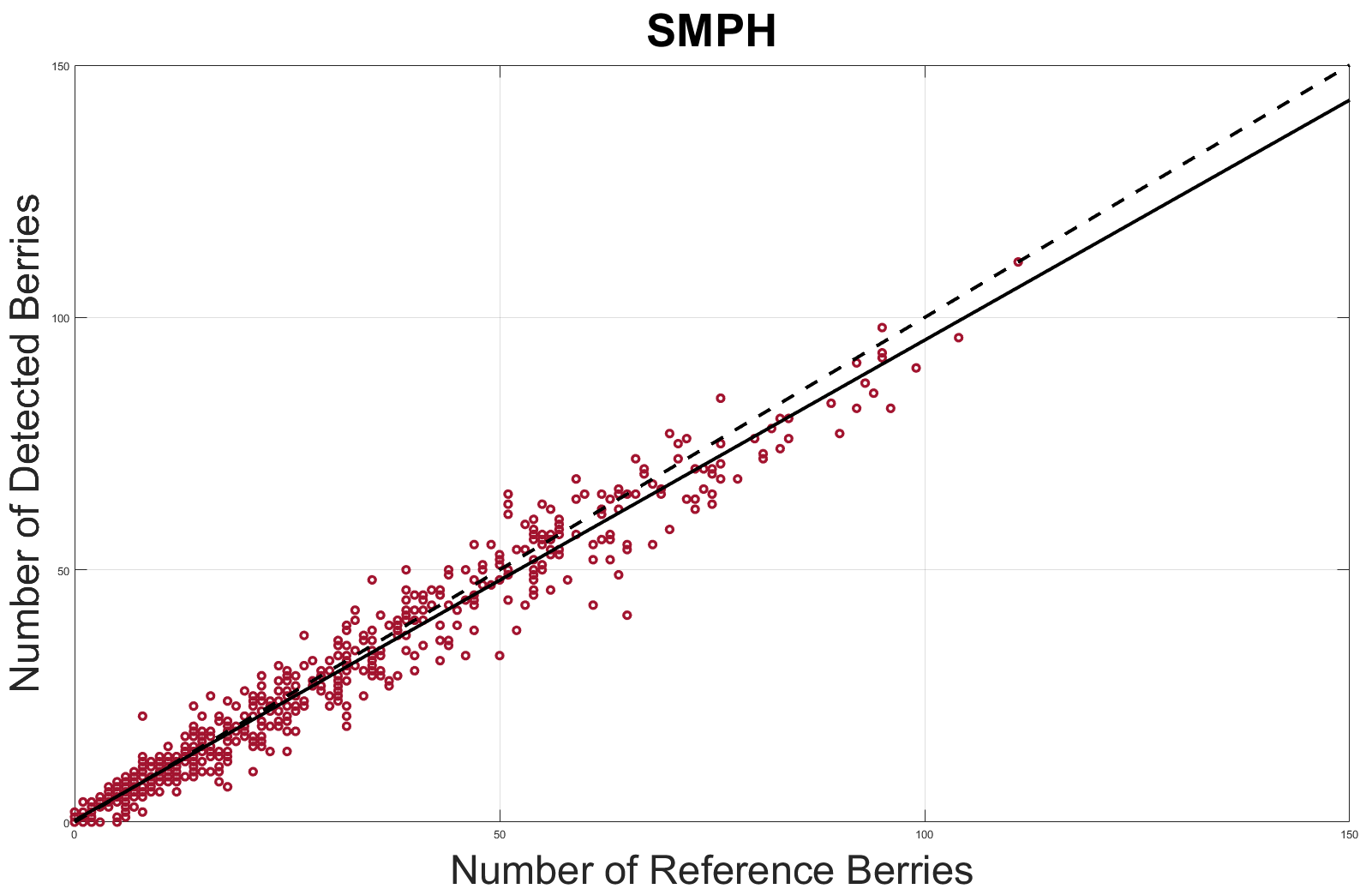}
        \caption[The classified image patches are used to reconstruct a mask for the original image using a majority vote.]%
        {{\small $R^2$-Plot for the SMPH with $R^2 = 93.14$}}    
        \label{fig:R2_SMPH}
    \end{subfigure}
    \caption[ The average and standard deviation of critical parameters ]
    {\small $R^2$ Plots for the VSP and SMPH. The red circles depict the berry count for an image patch. The dashed line represents the optimal mapping, the continuous line the estimated one.} 
    \label{fig:R2}
\end{figure*}

Fig. \ref{fig:R2} shows the $R^2$-plots for the VSP and SMPH. In general the correlation between the number of detected berries and manually counted berries is high. We tend to underestimate the number of berries slightly. The worse correlation in the VSP training system, which is indicated by the $R^2$ of 88.33 tends to undersegment the berries more because we have a compact bunch structure where berries overlap each other. When the underlying berries are seen only by a fraction we are sometimes not able to distinguish it from the berry above. 

In contrast to that we are able to distinguish single berries with a better confidence in the SMPH because the bunch structure is looser than in the SMPH. Therefore we are able to distinguish the single berries with a higher confidence ($R^2$ of 93.14).

\subsection{Discussion}
Examples for incorrectly classified objects occur at leaf borders or round structures. An example for a vine which growth in a circle is shown in Fig. \ref{fig:Filter}. 

\begin{figure}[t]
   \includegraphics[width=1\linewidth]{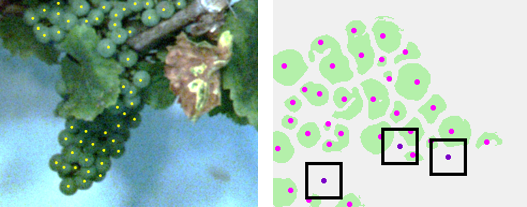}
   \caption{Detection problems. On the left side manually guessed berries in dark image regions cannot be found by network. On the right side is an example where multiple manually marked berries lie in the same component. The green areas with the pink dots show correctly detected berries. The dark dot in the middle of the square is a manually marked berry which was not detected by our network.}
\label{fig:problems}
\end{figure}

By removing objects which do not meet the geometric roundness assumption we are able to remove many of these misclassified objects.
Another problem is that the classifier tends to undersegment berries. This means, that more than one manually counted berry lies within one detected component, see Fig. \ref{fig:problems}.

After applying these simple filter methods we are able to reduce the misclassification to $10 - 15\%$ by decreasing the correct detections by not more than $2\%$.

\section{Conclusion}
Wine breeders and viticulture are especially interested in yield estimation and forecasting. The traditional procedures for this involve skilled experts which extrapolate small sampled and combine them with historic data and experience. This is subjective and error-prone. Therefore the application of sensor systems, for example cameras, offer a cheap, fast and objective alternative.

In this paper, we present a robust pipeline for the detection and counting of berries in images by using a semantic segmentation approach. With this we simplify the instance segmentation task.
We aim to detect single berries with the final goal of yield estimation and forecasting in mind. For future work and an even more sophisticated yield estimation we are able to provide even the berry size, although that was not investigated in this work.

We investigated the intersection over union (IoU) of the network on the test set. The class 'berry' is detected with a higher confidence (IoU = 76 \%). The class 'edge' is defined to be the boundary with a two pixels thickness. Therefore it is hard to detect the same edge precisely but an IoU of more than 40 \% is still satisfactory.

We evaluate our network on 60 independent and dot annotated images, divided into half showing to different training systems: vertical shoot positioned (VSP) and semi minimal pruned hedges (SMPH). We show that we can identify up to $87\%$ of berries in the leaf covered areas of the SMPH. For the VSP we are able to detect up to $94 \%$ of berries correctly.

\section*{Acknowledgment}
This work was partially funded by German Federal Ministry of Education and Research (BMBF, Bonn, Germany) in the framework of the project novisys (FKZ 031A349)

{\small
\bibliographystyle{ieee.bst}
\bibliography{egbib.bib}
}
\end{document}